\documentclass{svproc}
\usepackage{graphicx}
\usepackage{caption}
\usepackage{subcaption}
\usepackage{url}
\usepackage{textcomp}
\usepackage{mathtools} 
\usepackage{amsmath}
\usepackage{ragged2e}
\usepackage{latexsym}
\usepackage{xcolor}
\usepackage{threeparttable}

\usepackage[vlined, ruled, linesnumbered, commentsnumbered]{algorithm2e}
\usepackage{color}
\usepackage{dblfloatfix}
\usepackage{booktabs}

\begin{document}
\mainmatter              
\title{Arm Manipulation Planning of Tethered Tools with the Help of a Tool Balancer}
\titlerunning{Regrasp Planning with Tool Cable Constraints}  
%
\author{Daniel S\'anchez \and Weiwei Wan \and Kensuke Harada }
\authorrunning{D. S\'anchez, W. Wan, and K. Harada} 
%
\tocauthor{Daniel S\'anchez, Weiwei Wan, Kensuke Harada}
\institute{Osaka University, Japan.
\email{wan@hlab.sys.es.osaka-u.ac.jp}}

\maketitle              

\begin{abstract}
Robotic manipulation of tethered tools is widely seen in robotic work cells.
They may cause excess strain on the tool's cable or undesired entanglements with the robot's arms.
This paper presents a manipulation planner with cable orientation constraints for
tethered tools suspended by tool balancers. The planner uses orientation constraints
to limit the bending of the balancer's cable while the robot manipulates a tool
and places it in a desired pose. 
The constraints reduce entanglements and decrease the torque induced by the cable
on the robot joints. Simulation and real-world experiments show that the constrained
planner can successfully plan robot motions for the manipulation of suspended
tethered tools preventing the robot from damaging the cable or getting its arms
entangled, potentially avoiding accidents. The planner is expected to play
promising roles in manufacturing cells.
\keywords{Manipulation planning, Constraints, Wired tools}
\end{abstract}
\section{Introduction}

Manipulation planning involves the computation of robot motions to pick up objects
and place them in desired poses and requires important considerations to be successful:
The initial and goal poses of the manipulated object must be determined, the 
different grasping poses available to the robot in order to handle the object 
must be considered, and the intermediate poses of the robot for achieving the
object's goal pose must be computed and executed. Furthermore, planning constraints
are often implemented to guarantee that the manipulation task can be performed 
correctly by preventing unwanted events or following certain end-effector
poses \cite{wan2016developing,cheng1995line}. In 
industry-based scenarios, a robot usually has to handle tools to execute
different tasks. In those cases, the tool's inherent characteristics such
as its weight, its shape and (in the case of wired tools) its cable represent
hindrances for tool manipulation. A tool's cable, in particular, represents
a difficult problem for robotic manipulation: the robot might get entangled with
the cable, causing damage to the tool or itself or it might bend the cable in
excess causing wear and tear or its breakage. Also, the properties of the cable
itself, such as its fixed points and its flexibility play a major role in the
manipulation process. These burdens make the handling of wired or tethered tools
a challenge for most manipulation planners, therefore the inclusion of special
constraints to prevent or diminish entanglements and excessive cable bending is
of great importance for tethered tool manipulation. 

In this paper, we present a solution for manipulation planning of tethered tools
suspended by tool balancers. Tool balancers are suspended mechanisms placed in 
a workstation to help to fix tools. A tool balancer is equipped with a retractable 
cable that forces the tool's cable to form a straight line by
constantly applying a pulling force. The straight line simplifies
the problem of cable deformation. This work presents a set of constraints implemented 
with tool balancers to keep the robot from twisting the cable more than a given
maximum angle threshold. The planning algorithms under the constraints help to eliminate
robot motions that cause tethered tools' cables to bend excessively as well as 
avoid colliding with the cables before grasping, reducing entanglements and 
large strain. 

\begin{figure*}[!htbp]
\centering
 \includegraphics[width=.9\textwidth]{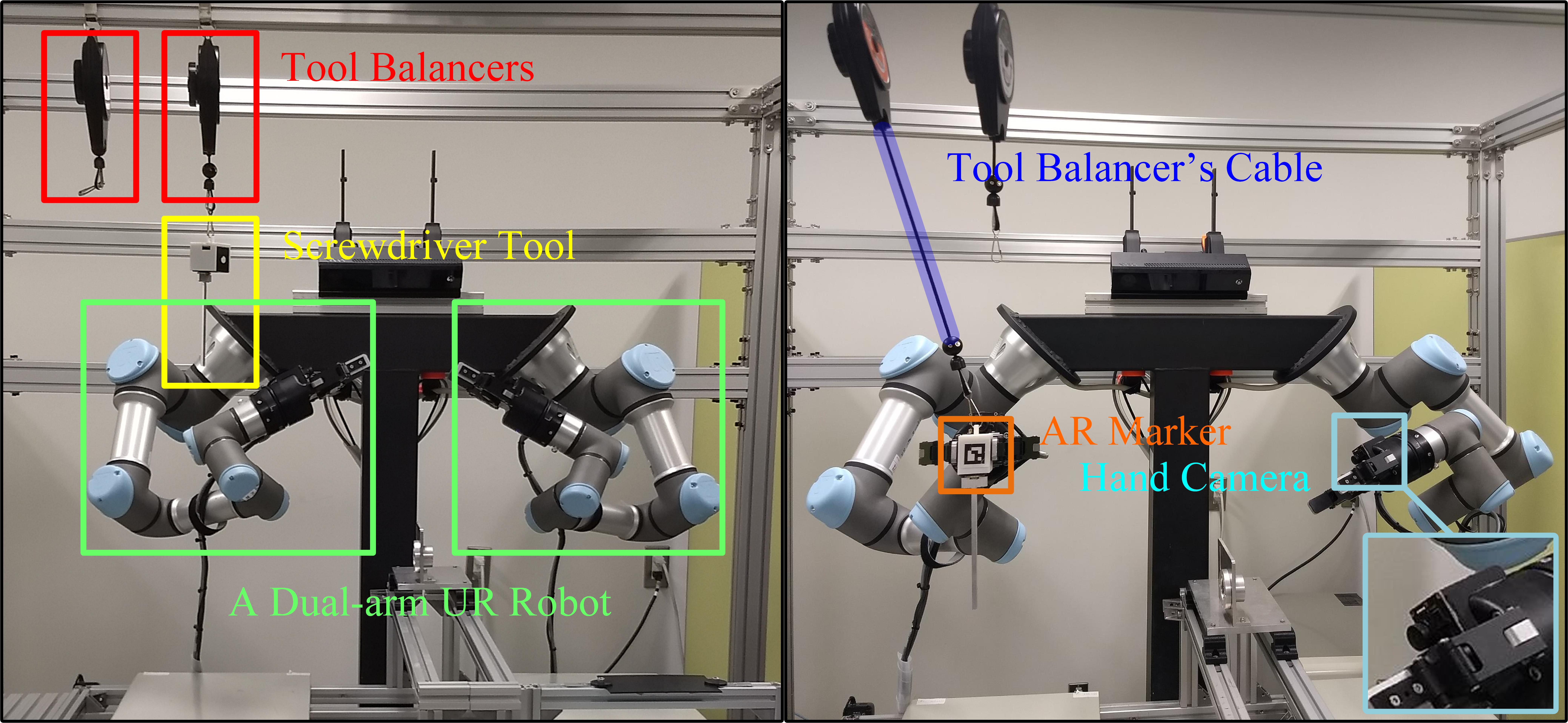}
 \caption{The real-work environment. Left: The tool balancers, the screwdriver
 tool, and the dual-arm UR3 robot (UR3D). Right: The tool balancer's cable, 
 the connection point between the tool and the cable, the AR marker used to determine the 
 pose of the tool and the camera mounted in a robot arm to detect the marker.}
 \label{fig:settingrealjoined}
\end{figure*}

To validate the proposed methods, we performed both simulations and real-world experiments
using the setting shown in Fig.\ref{fig:settingrealjoined} and a controller developed in
our previous work \cite{wan2016developing,wan2015reorientating}.
The results show a significant increase in manipulation success rate and a reduction of
the tool's cable bending as well as the torque borne by the robot arms.


\section{Related Work}

For decades, manipulation planning has captured the interest of several research 
projects. Manipulation consists in finding a series of intermediate object poses
in order to reach a goal pose. Early work in manipulation solutions for robotic 
arms \cite{tournassoud1987regrasping,stoeter1999planning,lozano1992handey} explored 
regrasp planning in fixed single-armed robots. These work presented some guidelines 
for future research in manipulation planning. Recently, multiple work on manipulation 
planning has been presented. These work includes the comparison of single-arm and 
dual-arm algorithms for regrasping \cite{wan2016developing,calandra2018more,wan2015reorientating}, 
a regrasp control policy that uses tactile sensors to plan adjustments for local 
grasp \cite{hogan2018tactile}, a planner that computes robot configurations to grasp 
assembly parts for sequences of collaborative assembly operations \cite{dogar2015multi} 
and 3D object reorientation using pivoting \cite{hou2018fast}. 

In the case of manipulation planning, the consideration of constraints is often 
used to ensure the correct execution of a manipulation task in restrictive 
scenarios: In \cite{perez2017c}, a method of learning multi-step manipulation was 
presented, it uses demonstration to teach the task as a sequence of key frames and 
a set of geometric constraints. In \cite{mirabel2017manipulation} a tool used to 
describe the various motion constraints relative to a manipulation
planning problem is proposed. A method for developing locally optimal trajectories 
for aerial manipulation in constrained environments is presented in \cite{seo2017locally}.
A solution for regrasp planning considering stability constraints in humanoid 
robots was shown in \cite{sanchez2018regrasp}. These approaches help to illustrate 
the usefulness of considering environmental or inner constraints in different 
manipulation scenarios. 

On the other hand, robot motion planning involving cable-like objects has 
presented many challenges, these objects are difficult to manipulate, with 
often unpredictable deformations and, in the case of cables, they can get 
entangled around the robot or its surroundings. Several work centered around 
manipulation planning have tackled different problems associated with deformable 
objects like cables, chains and hoses \cite{igarashi2010homotopic,hert1996ties,pham2018robotic,bretl2014quasi,ramirez2016motion}. 
In particular, deformable object manipulation has been explored in different 
work: In \cite{zhu2018dual} a framework for cable shapes manipulation using a 
deformation model was shown. A motion planner for manipulating deformable linear 
objects and tying knots is described in \cite{saha2006motion} and 
\cite{khalil2010dexterous} shows a review on dexterous robotic manipulation 
of deformable objects by using sensor feedback.

The aforementioned work on manipulation of deformable objects and constraints 
implementations suggested general solutions for object reorientation under 
different problematics, but the electric wires of tools are overlooked. It remains difficult
to include cable shape prediction, collision, and entanglement avoidance, 
and motion planning in the same loop. This paper uses a tool balancer to 
convert the cable constraints into a straight line constraint
and implements a planner that addresses the problems caused by wired cables using the constraints.
It helps to limit excessive bending of the cable, 
reduces stress on the robot joints, and help robots to avoid getting entangled with the cable and
hitting the tool's cable before performing a grasp.

\section{The Orientation Constraints}

The essence of our constraints is the estimation of the angular 
difference of a cable's current orientation vector and a tool balancer vector.
The cable's current orientation vector is represented as
a vector that points to the tool's connection point with the cable.
The tool balancer vector is represented as a vector 
that points to a desired cable orientation. The Cartesian angular difference 
between the two vectors is measured for each robot pose that the planner considers 
to build the motion path. If the measured angle is above a given threshold, the 
cable is considered to be excessively bent and the pose/node is discarded, the 
planner then continues exploring different robot poses until a set of nodes 
connecting the initial and goal poses of the object is found. The planner also 
generates a list of obstacles to represent the tool balancer's cable, and
prevents the robot from colliding with the cable before grasping the tool.

Specifically, the planner uses 
two unitary vectors defined in the world coordinate frame 
$\Sigma_{\text{o}}$, say $ {}^{\text{o}}_{}{\boldsymbol{\mathit{a}}}^{}_{}$ and
$ {}^{\text{o}}_{}{\boldsymbol{\mathit{b}}}^{}_{}$, to calculate the changes in 
angle between the cable orientation and the 
tool's current orientation.


First, our planner 
evaluates the orientation of a vector $ {}^{\text{t}}_{}{\boldsymbol{\mathit{b}}}^{}_{}$ 
in the tool's reference frame. This vector points to the connection point of the 
tool to the balancer's cable disregarding the tool orientation. Second, we use a 
vector $ {}^{\text{o}}_{}{\boldsymbol{\mathit{a}}}^{}_{}$ in the world frame to describe the vector
that points to the same connection point when the tool is in its initial suspended pose.
To transform them to the world frame, we use 
${}^{\text{o}}_{}{\boldsymbol{\mathit{b}}}^{}_{} = 
{}^{\text{o}}_{}\text{\textbf {R}}^{}_{\text{t}}  {}^{\text{t}}_{}{\boldsymbol{\mathit{b}}}^{}_{}$
where $ {}^{\text{o}}_{}\text{\textbf {R}}^{}_{\text{t}} $ is the object's 
rotational matrix. In the case of the 
tool balancer, we made $ {}^{\text{o}}_{}{\boldsymbol{\mathit{a}}}^{}_{} = {}^{\text{o}}_{}{\boldsymbol{\mathit{z}}}^{}_{}$ 
since the balancer is directly above of the tool. Initially the vector that points to the connection point
only has a ${}^{\text{o}}_{}{\boldsymbol{\mathit{z}}}^{}_{}$ 
component in the world reference frame.

\begin{figure*}[!h]
\centering
 \includegraphics[width=.9\textwidth]{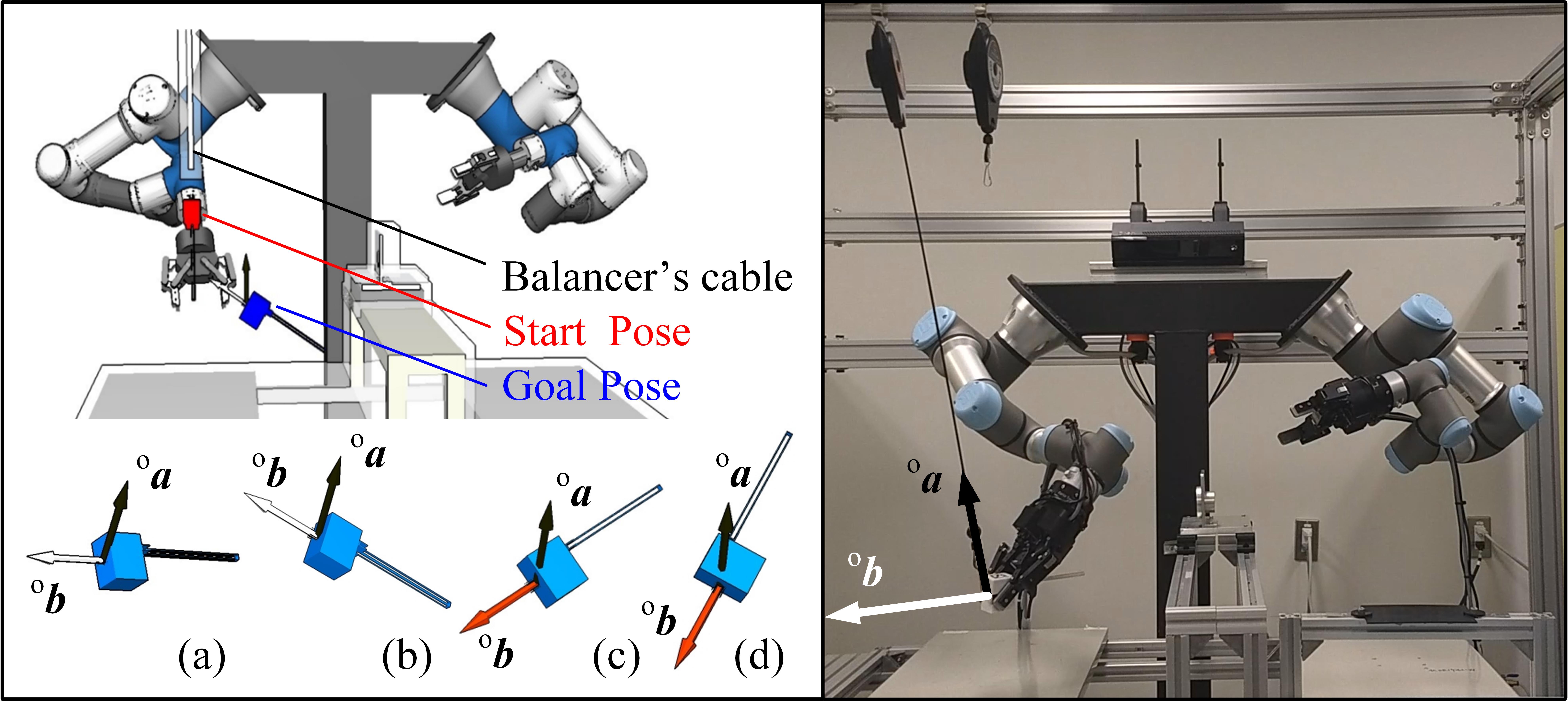}
 \caption{The simulation environment. (a-d) show the tool and the two 
 vectors used to determine the object's orientation: The black arrow represents 
 vector $ {}^{\text{o}}_{}{\boldsymbol{\mathit{a}}}^{}_{}$. If the angular 
 difference between ${}^{\text{o}}_{}{\boldsymbol{\mathit{a}}}^{}_{}$ and the
 direction of the connector
 ($ {}^{\text{o}}_{}{\boldsymbol{\mathit{b}}}^{}_{}$, as illustrated
 by the white arrows in (a, b) and the red arrows in (c, d)) 
 is greater than a given threshold (95\textdegree~in this case), the cable is 
 considered to be excessively (the red arrows indicate
 the angle has been violated).} 
 \label{fig:simulationsetting}
\end{figure*}

The angular difference between $ {}^{\text{o}}_{}{\boldsymbol{\mathit{a}}}^{}_{}$ and
$ {}^{\text{o}}_{}{\boldsymbol{\mathit{b}}}^{}_{}$ is computed using
$\theta$= $\arccos\frac{  {}^{o}_{}{\boldsymbol{\mathit{a}}}^{}_{}\cdot {}^{o}_{}{\boldsymbol{\mathit{b}}}^{}_{}}{
\mathopen|  {}^{o}_{}{\boldsymbol{\mathit{a}}}^{}_{}\mathclose|\mathopen| {}^{o}_{}{\boldsymbol{\mathit{b}}}^{}_{}\mathclose|}$.
If $\theta$ is higher than a given threshold the planner considers the cable to 
be excessively bent, discards the robot pose and continues searching the regrasp
graph for other grasps until a path is found. In Fig.\ref{fig:simulationsetting} 
a demonstration of our simulation environment with different tool orientations 
is shown to better illustrate vectors $ {}^{\text{o}}_{}{\boldsymbol{\mathit{a}}}^{}_{}$ 
and $ {}^{\text{o}}_{}{\boldsymbol{\mathit{b}}}^{}_{}$ and their angular difference.

We integrated the constraints into a planner published in
\cite{wan2016developing}. First, the planner determines the manipulated tool's 
position and orientation by using the robot's hand camera and the tool's 
AR marker, this pose corresponds to the objects starting pose, Second, 
the object's goal pose is given to the planner to start the planning process. 
Third, the cable is represented as an obstacle in the planner: The cable's shape 
can be approximated to a straight line, represented by a white cylinder in our 
simulator as seen in Fig.\ref{fig:settingrealjoined}. By defining this cylinder 
as an obstacle, we use collision detection to make the robot avoid the cable
before the it grasps the object. Finally, after
finding robot poses to grasp the object, the planner continues to build a motion 
sequence by sampling and connecting a series of compatible grasp poses, the 
poses must be IK-feasible, collision-free and must assure that the object 
orientation satisfies the orientation constraints. If the aforementioned $\theta$ is 
higher than a given threshold, the robot pose violates the constraints and is discarded. 
In this way, the 
planner builds a motion sequence for the robot to grasp the object and 
place it in a desired goal pose while preventing excessive bending of the 
cable, avoiding possible entanglements and reducing the strain applied on 
the robot by the retractable cable of the tool balancer.

\section{Experiments}
To validate the implementation of the aforementioned constraints, we performed both
simulation and real-world experiments.
For simulations, the tests involved the 
manipulation of a 3D mesh model of our screwdriver tool. The robot was tasked
to pick the tool from a hanging position and place it in a desired goal pose,
the experiment was conducted several times with slight variations in the roll
and pitch of the starting and goal poses of the tool to test different scenarios. 
Comparisons were performed with the constrained and unconstrained planner.
In the real-world experiments, we used two UR3 arms with mounted cameras in each robot gripper, 
the cameras were used to capture the original position and rotation of the 
screwdriver tool with the aid of AR markers. The robot is asked to perform several tasks with 
distinct goal poses. The robot's torque sensors and current sensors were used
to measure the strain produced by the tool balancer's cable on the robot.

\subsection{Simulations}

We performed a series of tests in our simulation environment, shown
in Fig.\ref{fig:simulationsetting}, to measure and compare the planner's success rate 
with different manipulation tasks. We labeled the execution of the tasks using 
the following criterion: If the tool's angle difference $\theta$ is higher
than a given threshold angle at any point of the task, the execution is labeled as failed by excessive 
bending of the cable; If the robot hits the balancer's cable before grasping 
the object, the execution is identified as a failed case due to collision (In 
real-world experiments this would modify the tool's starting pose, making 
the planned motion ineffective); If the planner is not able to find a path 
to complete the task, the execution is also labeled as a failure; If 
the robot is able to perform the task without violating any of the constraints,
the execution is labeled as success. Both
constrained and unconstrained planners are simulated and labeled following the criterion.

\begin{figure*}[!t]
\centering
 \includegraphics[width=.98\textwidth]{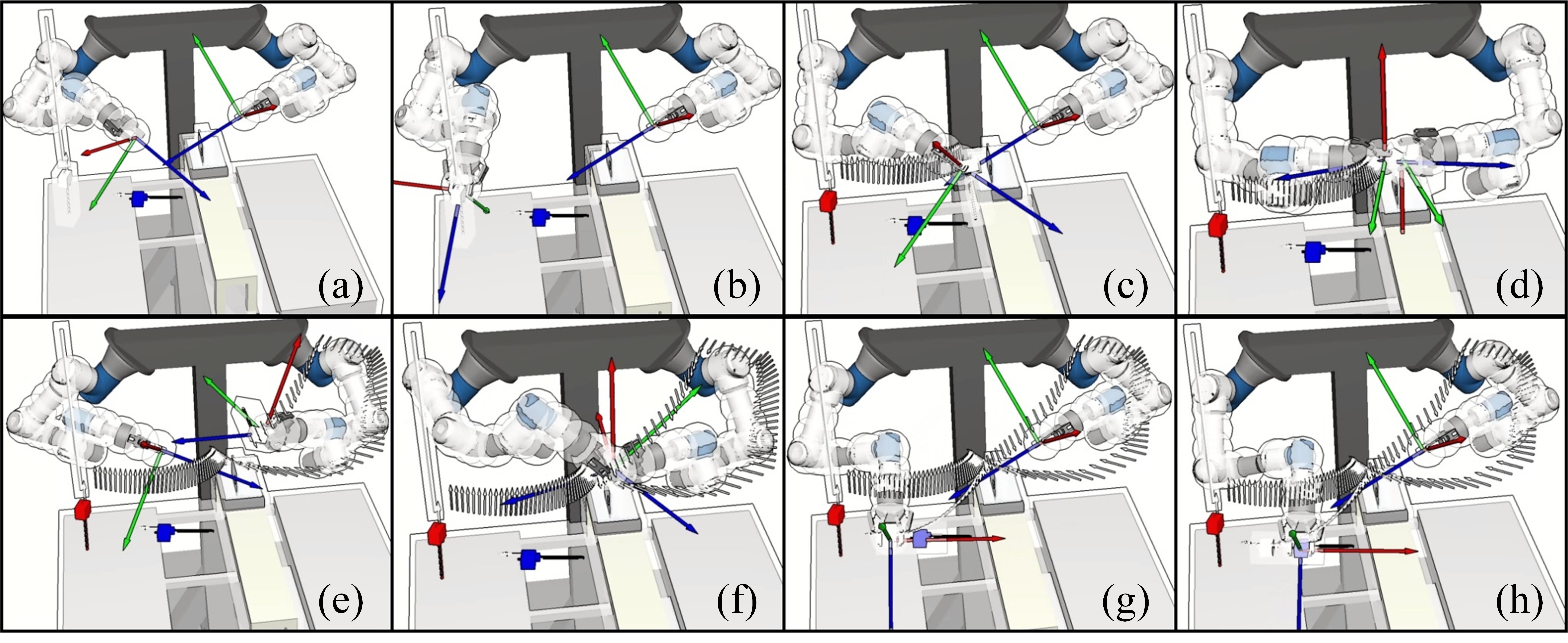}
 \caption{A successful sequence planned in the simulator. 
 (a, b) The robot reaches to the screwdriver in order to grasp it. 
 (c, d) it moves the tool closer to its left hand in order to perform a handover. 
 (e, d, f) The robot re-positions the tool with its left arm without incurring into undesired 
 tool orientations. The white arrows show the position and orientation of 
 $ {}^{\text{o}}_{}{\boldsymbol{\mathit{b}}}^{}_{}$. 
 (g, h) The robot regrasps the tool with its right arm to move it to the goal.} 
 \label{fig:experiments2constrainedjoined}
\end{figure*}

We tasked both constrained and unconstrained planners to find a motion for the 
robot to grasp the object from a starting pose and move it to a goal pose. 
We varied the combinations of starting and goal poses by introducing 
roll and pitch angular rotations (in the tool's frame of reference $\Sigma_{\text{t}}$). 
The results are shown in Table \ref{Tab: Simulation results for the constrained and unconstrained planner}.
The columns and rows in Table \ref{Tab: Simulation results for the constrained and unconstrained planner}
are the varied starting and goal poses combinations.
In total, we introduced 5 different goal poses (the columns of 
Table \ref{Tab: Simulation results for the constrained and unconstrained planner}).
For each one of these goals, we find the motion for 8 different starting poses
(the rows of Table \ref{Tab: Simulation results for the constrained and unconstrained planner}).
We performed a total of $5\times8\times2$=$80$ tests.
For the constrained planner, the angle $\theta$ is required to be 
smaller than 95\textdegree at any point in the motion, 
otherwise the execution will be labeled as a failure. An example of a successful 
execution considering the constraints is shown in Fig.\ref{fig:experiments2constrainedjoined}. 

\begin{table}[!htbp]
\centering
\setlength{\tabcolsep}{7.3pt}
\caption{\label{Tab: Simulation results for the constrained and 
unconstrained planner}Simulation results for the constrained and unconstrained planner}

\begin{tabular}{cccccccccccccc}
\toprule
&\multicolumn{5}{c}{Constrained}  & &  & \multicolumn{5}{c}{Unconstrained}\\
\cmidrule{2-6} \cmidrule{9-13}
&-20\textdegree& -10\textdegree & 0\textdegree & 10\textdegree &  20\textdegree & &&-20\textdegree& -10\textdegree & 0\textdegree & 10\textdegree &  20\textdegree
 \\
 \midrule
 \midrule
  0\textdegree & $\circ $ &$\circ $ &$\circ $&$\circ $ & $\circ $& &  & $\circ$ &$\circ$ &$\circ$ &$\circ$ & $\star$& \\
 \midrule
  10\textdegree& $\circ$ &$\circ$ &$\circ$ &$\circ$ & $\circ$&  & & $\circ$ &$\circ$ &$\circ$ &$\star$ & $\circ$& \\
\midrule
  15\textdegree & $\circ$ &$\circ$ &$\circ$ &$\circ$ & $\circ$& & & $\times$&$\circ$ &$\circ$&$\circ$ & $\circ$& \\
\midrule
 30\textdegree& $\circ$ &$\circ$ &$\circ$ &$\circ$ & $\circ$&  &  & $\times$&$\circ$ &$\circ$ &$\times$ & $\circ$&\\
\midrule
  45\textdegree & $\circ$ &$\circ$ &$\circ$ &$\circ$ & $\circ$& & & $\star$&$\times$ &$\circ$&$\times$ & $\times$&  \\
\midrule
 60\textdegree& $\circ$ &$\circ$ &$\circ$ &$\circ$ & $\circ$& &&  $\circ$&$\circ$ &$\star$&$\times$& $\circ$& \\
\midrule
  75\textdegree &$\bigotimes$&$\bigotimes$ &$\circ$&$\bigotimes$ & $\bigotimes$& & &$\times$&$\circ$& $\times$ &$\times$ &$\times$&\\
\midrule
  90\textdegree &$\bigotimes$&$\bigotimes$ &$\bigotimes$ &$\bigotimes$ & $\bigotimes$& & &$\times$&$\circ$&$\circ$ &$\bigotimes$ & $\times$&\\
\midrule
\bottomrule
\end{tabular}
\begin{tablenotes}
\justifying{
\item[Meanings of abbreviations] Results of the constrained and unconstrained planner. 
The angles in columns and rows show the roll and pitch rotations (in the tool's frame of reference 
$\Sigma_{t}$) of the starting and goal poses. 
Symbols: $\circ$ indicates successful planning; 
$\times$ indicates that the cable was bent above the threshold angle 
of 95\textdegree; $\star$ indicates collisions with the balancer's cable before
grasping the object. $\bigotimes$ indicates failure cases where the planner 
could not find a motion.}
\end{tablenotes}
\end{table}

The simulation results yielded a 77.5\% success rate for the constrained planner. The planner was 
not able to compute viable solutions for the cases were the roll rotation 
was of 90\textdegree, most likely because the unconstrained solution space 
for that particular configuration is very limited, making the bending of the 
cable a requirement to reach the goal pose. In the case of the unconstrained planner, 
the success rate derived from Table \ref{Tab: Simulation results for the constrained 
and unconstrained planner} (counting cable collisions, and cable bending as failures) 
was of 57.5\%. In some cases the robot would hit the cable in the simulation (this 
does not modify the tool's starting pose in the simulation unlike the real-world) 
and also bend the cable in order to complete its task, in said cases the results 
were labeled as failures by bending of the cable.

\subsection{Real-world experiments}
\begin{figure*}[h]
\centering
 \includegraphics[width=.9\textwidth]{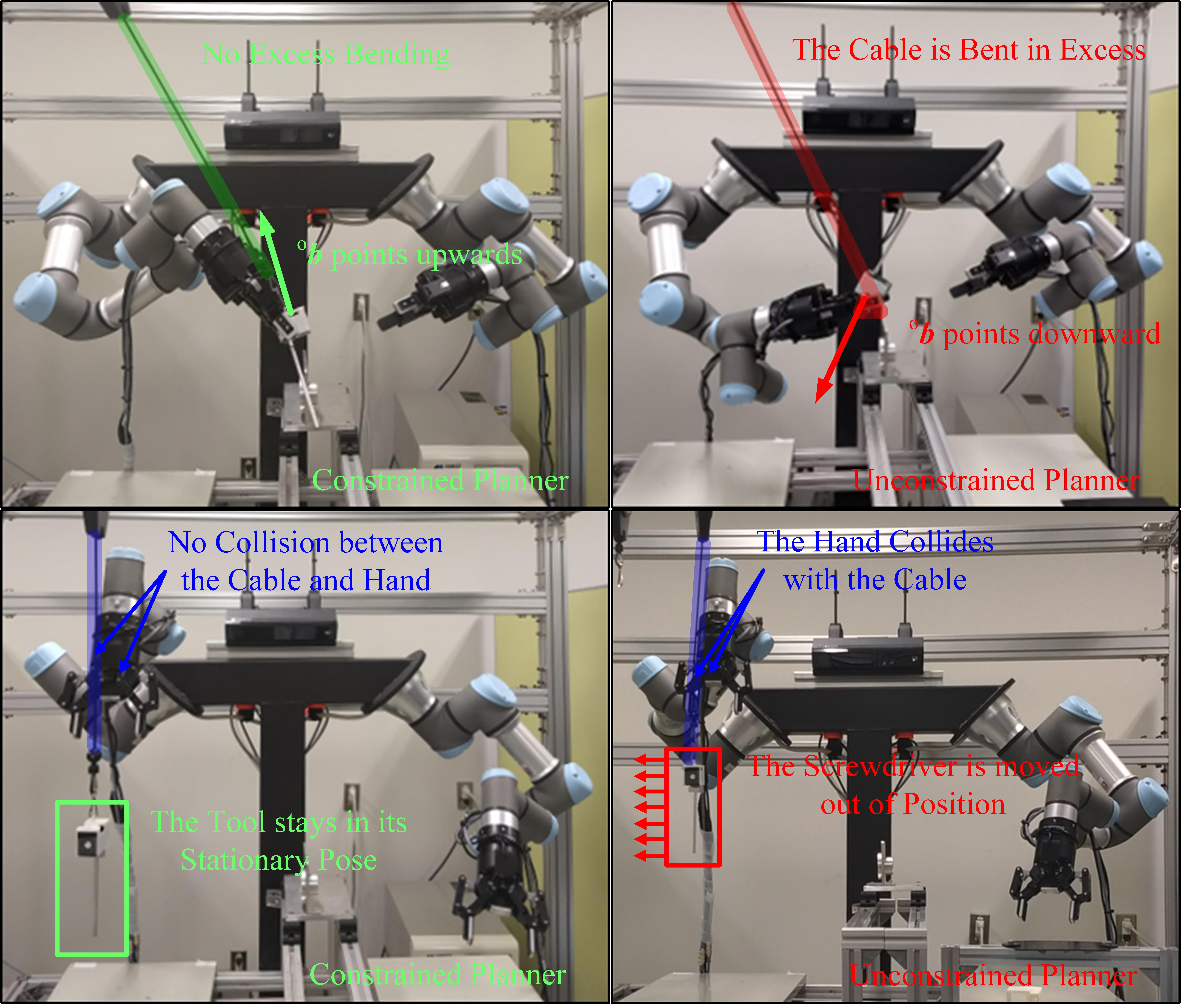}
 \caption{Comparison between the constrained (pictures on the left) and 
 unconstrained (pictures on the right) planner. With the constraints in 
 place the robot is able to successfully complete its task without bending 
 in excess the cable or colliding with it.} 
 \label{fig:collagebendandcollision}
\end{figure*}

For real-world experiments, we used the dual-arm UR3 robot and a tool 
balancer to suspend a small screwdriver tool, the robot uses hand-mounted 
cameras an AR markers to identify the starting pose of the object, as seen 
in Fig.\ref{fig:settingrealjoined}. The experiments were performed with and 
without the constraints presented in this work in order to compare their 
performances. We tasked the robot to place the tool in several goal poses. 
After performing several experiments we found that the constrained planner 
successfully avoids the two undesired cases: When the robot bends the cable 
in excess or when it collides with it. Fig.\ref{fig:collagebendandcollision} 
better illustrates both cases and the solutions presented by the unconstrained 
planner to avoid them. Statistically, real-world experiments with the 
constraints yielded a 80\% success rate for 15 different executions. 
In the 3 failed cases, the planner was not able to find a solution.
On the other hand, the success 
rate of the unconstrained planner for the same 15 executions was 46.6\%. 
5 of the failure cases were caused by collisions with the cable before 
grasping the tool. The other 4 failure cases were caused by excessive cable bending. 
A video comparing the constrained 
and unconstrained executions is available at: 
http://y2u.be/uTIeeQMYwJ4.

\begin{figure*}[h]
\centering
 \includegraphics[width=.8\textwidth]{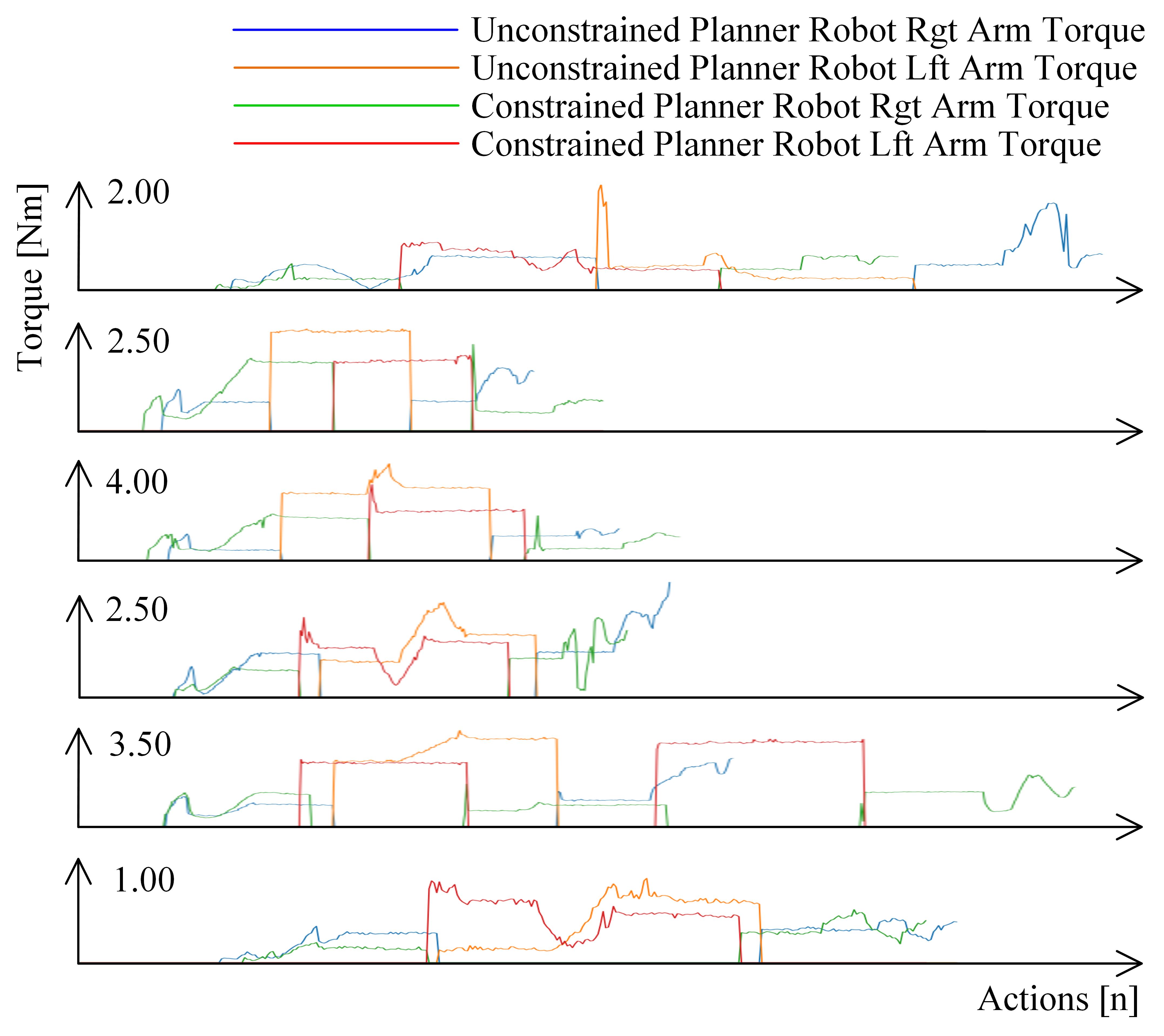}
 \caption{Comparison between the constrained and unconstrained planner. 
 With the constraints, the total torque resisted by the robot due to the
  torque of the balancer's cable is reduced. The first four pictures show results with a  balancer of 2 Kg maximum 
  load tool, while the last two figures illustrate experiments performed with a balancer 
  of 0.8Kg of maximum load. 
  Experimental results show that the torque applied by the cable is lower when 
  the planner the proposed our constraints, effectively reducing the strain on the robot joints. } 
 \label{fig:realexperimentstorqueresults}
\end{figure*}

On the other hand, we used torque sensors to measure the changes of strain on the robot's joints. 
Fig. \ref{fig:realexperimentstorqueresults} shows the magnitude values. The torque magnitudes were 
recorded every time one of the arms grasps the object, the strong spikes 
in the measurement occurred when the robot performed handover operations 
between its grippers. The results show that our constraints reduce the 
magnitude of the torque suffered by the robot. On average the constrained 
planner reduced the maximum torque of the two arms by
26.1 \% and 20.5 \% respectively, diminishing the strain on the 
robot and the balancer's cable.

\section{Conclusions}

A constrained planner for suspended tethered tool manipulation was presented.
The planner is examined by both simulation and real-world experiments.
Results confirmed that the planner can successfully compute  motion sequences for robots to handle 
balancer-suspended tools, preventing collisions with s tool's cable and its 
excess bending, reducing the strain suffered by the robot by 
diminishing the torque applied by the balancer's retractable cable on its 
joints. The planner is expected to play promising roles in manufacturing cells.

\bibliographystyle{spmpsci.bst}
\bibliography{BibDaniel}

\end{document}